\documentclass{article}

\usepackage[final]{neurips_2023}

\usepackage[utf8]{inputenc} % allow utf-8 input
\usepackage[T1]{fontenc}    % use 8-bit T1 fonts
\usepackage{hyperref}       % hyperlinks
\usepackage{url}            % simple URL typesetting
\usepackage{booktabs}       % professional-quality tables
\usepackage{amsfonts}       % blackboard math symbols
\usepackage{nicefrac}       % compact symbols for 1/2, etc.
\usepackage{microtype}      % microtypography
\usepackage{xcolor}         % colors
\usepackage{lmodern}

\usepackage{graphicx}
\usepackage{amsthm}
\usepackage{amsmath}
\usepackage{tikz}
\usepackage{tikz-qtree}
\usepackage{blkarray}
\usetikzlibrary{shapes,decorations,arrows,calc,arrows.meta,fit,positioning}
\tikzset{
    -Latex,auto,node distance =1 cm and 1 cm,semithick,
    state/.style ={ellipse, draw, minimum width = 0.7 cm},
    point/.style = {circle, draw, inner sep=0.04cm,node contents={}},
    empoint/.style = {, draw, inner sep=0.04cm,node contents={}},
    bidirected/.style={Latex-Latex,dashed},
    el/.style = {inner sep=2pt, align=left, sloped}
}

\title{Inherent Inconsistencies of Feature Importance}

\author{%
          Nimrod Harel\\
          Department of Computer Science\\
          Tel-Aviv University\\
          Tel Aviv, Israel \\
          \texttt{nimrodharel@mail.tau.ac.il} \\
           \And
           Uri Obolski\\
          School of Public Health\\
          Porter School of the Environment and Earth Sciences \\
          Tel-Aviv University\\
          Tel Aviv, Israel \\
          \texttt{uriobols@tauex.tau.ac.il} 
          \And
          Ran Gilad-Bachrach\\
          Department of Biomedical Engineering\\
          Tel-Aviv University\\
          Tel Aviv, Israel \\
          \texttt{rgb@tauex.tau.ac.il} \\
}

\begin{document}

\newtheorem{statement}{Statement}[section]

% \usepackage{tikz}
% \usetikzlibrary{positioning}    
% \usepackage{amsmath}
% \usepackage{threeparttable}
% \usepackage{natbib}
\setcitestyle{numbers}

%Addition
% \usepackage{float}
\newcommand\fnote[1]{\captionsetup{font=small}\caption*{#1}}
%\usepackage{subcaption}
% \usepackage{graphicx, nicefrac}
% \usepackage{algorithm}
% \usepackage{algpseudocode}
% \usepackage{tikz-qtree}
% \usepackage{blkarray}
% \usetikzlibrary{shapes,decorations,arrows,calc,arrows.meta,fit,positioning}
% \tikzset{
    % -Latex,auto,node distance =1 cm and 1 cm,semithick,
    % state/.style ={ellipse, draw, minimum width = 0.7 cm},
    % point/.style = {circle, draw, inner sep=0.04cm,node contents={}},
    % empoint/.style = {, draw, inner sep=0.04cm,node contents={}},
    % bidirected/.style={Latex-Latex,dashed},
    % el/.style = {inner sep=2pt, align=left, sloped}
% }

% \DeclareMathAlphabet{\pazocal}{OMS}{zplm}{m}{n}
\newcommand{\unif}{\pazocal{U}}
\newtheorem{property}{Property}
\newtheorem{theorem}{Theorem}
\newtheorem{lemma}{Lemma}
\newtheorem{definition}{Definition}
\renewenvironment{proof}{{\bf \emph{Proof.} }}{\hfill $\Box$ \\} 
\newenvironment{counterexample}{{\bf \emph{Counter-example.} }}{\hfill $\Box$ \\} 
\newenvironment{proofsketch}{{\bf \emph{Proof sketch.} }}{\hfill $\Box$ \\}

\newenvironment{nonumdef}{\begin{definition}\renewcommand{\thetheorem}{}}{\end{definition}}

\newcommand{\local}{l}
\newcommand{\dataset}{ \mathcal{X}}
\newcommand{\target}{ \mathcal{Y}}
\newcommand{\feats}{ \mathcal{F}}
\newcommand{\feat}{ f}
\newcommand{\groups}{ \mathcal{G}}
\newcommand{\group}{g}
\newcommand{\model}{ \mathcal{M}}
\newcommand{\data}{ \mathcal{D}}
\newcommand{\subgroup}{\mathcal{S}}
\newcommand{\reals}{\mathbb{R}}
\newcommand{\expected}{\mathbb{E}}

\newcommand{\val}{\nu}
\newcommand{\vald}{\val^\data}
\newcommand{\valm}{\val^\model}
\newcommand{\valg}{\val_\glob}
\newcommand{\valgd}{\valg^\data}
\newcommand{\vall}{\val_\local}
\newcommand{\valld}{\vall^\data}
\newcommand{\fif}{\varphi}
\newcommand{\fiff}{\Phi}
\newcommand{\fifg}{\varphi_\glob}
\newcommand{\fifl}{\varphi_\local}
\newcommand{\locglob}{\{\vall, \fifl,\valg,\fifg\}}
\newcommand{\locglobd}{\{\valld, \fifl,\valgd,\fifg\}}
\newcommand{\datamodel}{\{\val^\data, \fif^\text{\tiny data},\val^\model, \fif^\text{\tiny model} \}}
\newcommand{\loc}{\text{\tiny local}}
\newcommand{\glob}{\text{\tiny global}}
\newcommand{\fifdiag}{\fif(\dataset, \val(\dataset))}
\newcommand{\fifdiagloc}{\fif(x, \val(x))}

\newcommand{\rgb}[1]{\textcolor{cyan}{RGB:#1}}
\newcommand{\nh}[1]{\textcolor{blue}{NH:#1}}
\newcommand{\uo}[1]{\textcolor{brown}{UO:#1}}
\newcommand{\tcent}[1]{
  {\renewcommand{\arraystretch}{1.6}
\begin{tabular}{l} #1 \end{tabular}}}

\newcommand{\prop}{\emph}

\maketitle

\begin{abstract}
The rapid advancement and widespread adoption of machine learning-driven technologies have underscored the practical and ethical need for creating interpretable artificial intelligence systems. Feature importance, a method that assigns scores to the contribution of individual features on prediction outcomes, seeks to bridge this gap as a tool for enhancing human comprehension of these systems. Feature importance serves as an explanation of predictions in diverse contexts, whether by providing a global interpretation of a phenomenon across the entire dataset or by offering a localized explanation for the outcome of a specific data point. Furthermore, feature importance is being used both for explaining models and for identifying plausible causal relations in the data, independently from the model. However, it is worth noting that these various contexts have traditionally been explored in isolation, with limited theoretical foundations.

This paper presents an axiomatic framework designed to establish coherent relationships among the different contexts of feature importance scores. Notably, our work unveils a surprising conclusion: when we combine the proposed properties with those previously outlined in the literature, we demonstrate the existence of an inconsistency. This inconsistency highlights that certain essential properties of feature importance scores cannot coexist harmoniously within a single framework.
\end{abstract}

\section{Introduction}

\emph{Feature Importance} scores gauge the contribution of each feature to an outcome of a model.
% Most model-agnostic feature importance scores use a two-step process such that in the first step value is assigned to subsets of the features and in the second step, the score of individual features is derived from the values of subsets. 
Most model-agnostic feature importance scores use a two-step process: in the first step, value is assigned to subsets of the features. In the second step, the score of individual features is derived from the values of subsets. This two-step process allows for a discussion about the expected behavior of the value function and the feature importance score. Many feature importance scores have been proposed in the literature: the bivariate-association~\citep{covert2020understanding} evaluates a feature's importance based on its conditional attributes, independent of other features, ablation-studies~\citep{casagrande1974ablation, bengtson-roth-2008-understanding,hessel2018rainbow} quantify a feature's significance by assessing its contribution when removed from the entire feature set, SHAP~\citep{lundberg2017unified} computes feature importance as the mean of its contributions across various subsets of features, and MCI~\citep{catav2021marginal} determines importance as the maximal contribution among all possible feature subsets (see Table~\ref{tab:funcfamily}). SHAP and MCI use an axiomatic approach in which the expected behaviors are defined as properties, and the functions are derived to satisfy these properties. 

Feature importance scores can be categorized by two main attributes: the scope, i.e. \emph{local} vs \emph{global}, and the objective, i.e. \emph{data} vs \emph{model}. 
Methods focusing on local interpretations seek to explain individual predictions (e.g., the role of each feature in a patient's diagnosis~\citep{kulesza2015principles}). Conversely, methods focusing on global interpretation try to understand how each feature affects a phenomenon (e.g., the role of each gene in a particular disease~\citep{tan2018learning, fontana2013estimation}). Along the second axis, the data and the model are distinguished by the type of conclusion required. The objective of explaining the data is to infer conclusions about the world that are encoded in the data, as the scientist does in his research~\citep{danaee2017deep, mckinney2006machine, haljas2018bivariate}. The objective of explaining the model, however, is to use an explanation to monitor and debug a model, to ensure it is working as intended (e.g., as the engineer does for security purposes~\citep{karlavs2022data, guo2018lemna}).

Table~\ref{tab:2x2} maps feature importance research according to the local vs. global and data vs. model settings. Most feature importance scores thus far have focused on explaining models, although the data scenario has also been gaining increased attention in recent years. However, the quadrant of the data-local setting is still unexplored in the field of explainable AI. Perhaps this is due to the challenge of providing an accurate explanation as to why a specific outcome (rather than an average result) came into being (rather than being calculated by a model). For example, which characteristic of John Doe is responsible for the fact that he did, or did not, suffer a stroke? These types of questions pertain to individual causal effects that are notoriously difficult to estimate~\citep{pearl2009causality, hernan2010causal}.

\begin{table}
    \caption{Examples of common feature importance scores. $\fif$ denotes the importance function, which takes $\val$, the value function, and a feature $\feat\in \feats$ as inputs, and assigns an importance score.}
  \label{tab:funcfamily}
  \centering
  \begin{tabular}{|c c|}
%   {|p{2.5cm}|p{7cm}|}
  \hline
        \tcent{\bf{Name}} & \tcent{ \bf{Feature importance score: }$\boldsymbol{ \fif(\val, \feat)} $}\\
        \hline
         \tcent{Bivariate} & \tcent{$ \val(\{\feat\})$}\\
             \hline 
         \tcent{Ablation} & \tcent{$\val(\feats) -\val(\feats \setminus \{\feat\})$} \\
             \hline
         \tcent{Shapley} & \tcent{$\sum_{\subgroup\subseteq \feats \setminus \{\feat\}} \frac{|\subgroup|!(|\feats|-|\subgroup|-1)!}{|\feats|!}\cdot (\val(\subgroup\cup \{\feat\}) - \val(\subgroup))$} \\
            \hline
          \tcent{MCI} & \tcent{$\max_{\subgroup \subseteq \feats \setminus \{\feat\}} (\val(\subgroup \cup \{\feat\}) - \val(\subgroup))$} \\
             \hline
  \end{tabular}

\end{table}

Several studies have examined the relations between the different settings:
\citet{lundberg2020local} presented a global score that is computed by combining local scores, hence indicating that at least the local and the global settings are not independent. \citet{covert2020understanding} proposed a method of assigning global importance to features, which draws a connection with the local feature importance score of $\emph{SHAP}$~\citep{lundberg2017unified}.   \citet{chen2020true} defined distinctions between the data and the model and argued that the nature of an explanation depends on what one seeks to explain -- the data or the model. Nevertheless, most studies focus solely on one setting. The studies that do consider multiple settings, often do not present an explicit set of expectations for the relations between importance scores under the different settings.

In this work, we establish the expected behavior of feature importance scores across diverse contexts. Our objective is to formalize a set of properties that capture the anticipated consistency between local and global interpretations, as well as the alignment between data-driven and model-based assessments. An intriguing extension of this framework is the introduction of the data-local scenario, which, in theory, can be achieved by integrating our properties with axiomatic methods, wherein expected behaviors are rigorously defined, and functions are derived accordingly. In the global-data scenario, we employ a set of properties introduced by~\citet{catav2021marginal}, implying the MCI importance score. Leveraging our proposed local-global relation, one can derive the expected data-local importance score. Conversely, in the local-model scenario, \citet{lundberg2017unified} present the SHAP importance score as the only function that satisfies their proposed set of properties. Here, utilizing our proposed data-model relation, one can similarly obtain the expected data-local importance score.  However, to our surprise, in both cases, even a modest set of requirements leads to contradictions. 
% To overcome this challenge, we developed a method for grouping features together based on their values, which was able to avoid some of the obstacles we encountered. 
% \usepackage[left=2cm, right=2cm, top=0.2cm, bottom=2cm]{geometry}

\begin{table}[t]
  \caption{Examples of feature importance scores and their categorization according to the global/local and data/model settings.}
  \label{tab:2x2}
  \centering
  \begin{tabular}{|l|p{5.cm}|p{5.cm}|}
  \hline & \tcent{\bf{Global}}     & \bf{Local} \\
    \hline
    \tcent{\bf{Model}} & 
    % \begin{flushleft}   
    Additive-Importance-Measures~\citep{covert2020understanding}
    Bivariate-Association~\citep{covert2020understanding}
    % \hspace{10mm} 
    % \hspace{25mm}
    Ablation-Studies~\citep{casagrande1974ablation, bengtson-roth-2008-understanding,hessel2018rainbow} 
    FIRM~\citep{zien2009feature} 
 Tree-Shap~\citep{lundberg2020local}
 % \end{flushleft}
 &  
 % \begin{flushleft} 
 SHAP~\citep{lundberg2017unified} Lime~\citep{ribeiro2016should}  
    Gradient-Based-Localization~\citep{selvaraju2017grad}
    Relevance-Propagation~\citep{bach2015pixel}
    \hspace{10mm}
    TreeExplainer~\citep{lundberg2020local}
    % \end{flushleft}   
    \\
    \hline
    \tcent{\bf{Data}}     & 
    % \begin{flushleft}
    True-To-Data~\citep{chen2020true} MCI~\citep{catav2021marginal} UMFI~\citep{Janssen2022ultra}
    % \end{flushleft} 
    &      \\
    \hline
  \end{tabular}
\end{table}

The paper is structured as follows: Section~\ref{sec:setting} consists of a formulation of the framework that generalizes the two-step process of feature importance to all settings. In Section~\ref{sec:localglobal} we focus on the local-global consistency: we present its properties and then demonstrate that they are incompatible with a previous result that defined the data-global setting. At the end of the section, we provide a brief discussion of the nature of the contradiction. Section~\ref{sec:data model} follows a similar structure as the latter, except addressing the case of the data-model consistency, which contradicts a previous result that defined the model-local setting. 
Due to space constraints, we provide supporting proofs for our claims in the Appendix, along with an extended version of Theorem~\ref{theorem:incon1} that allows a clear demarcation between local and global importance scores. 
% Furthermore, we expound on a method referred to as 'separable sets,' which mitigates the challenges encountered earlier, ensuring that all importance functions listed in Table~\ref{tab:funcfamily} converge to a consistent outcome.
% In the last section, we focus on separable sets, a special case in which the difficulties previously encountered do not exist, and all the importance functions in Table~\ref{tab:funcfamily} converge to the same result. 
% We prove the existence of a unique and maximal such partitioning.

This work makes two key contributions: (1) We introduce a unified axiomatic framework that encompasses feature importance analysis in diverse settings, including global vs. local and model vs. data contexts.~ (2) We rigorously demonstrate inconsistencies within these settings, shedding light on disparities between global and local interpretations and between model-based and data-centric evaluations. These findings enhance our understanding of the nuances and challenges in the theory of feature importance analysis within machine learning interpretability.
    % (3) We propose a method to overcome inconsistencies by assigning scores to \emph{separable sets} of features instead of individual features.

\section{Framework}
\label{sec:setting}
We begin by introducing some notation: the setting consists of an input space $\dataset$, an output space $\target$, so that given a pair $(x,y) \sim (\dataset \times \target)$, the learning task is to predict $y$ by observing $x$.
Without loss of generality, $\dataset\subseteq\reals^{|\feats|}$, where $\feats$ is the set of available features. The explanation task is aimed to assign a score to each feature, based on its contribution to prediction. It consists of a two-step process: a \textbf{value function} is a function $\val: \{x, \dataset\}\times 2^{\feats} \rightarrow \reals$ that assigns a scalar to each subset of features $\subgroup \subseteq \feats$, where $\val(x, \subgroup)$ denotes the local value of a subset of features $\subgroup$ for a given pair $(x,y)$ and $\val(\dataset,\subgroup)$ denotes the global value of this subset. 
A \textbf{feature importance function} is a function 
$\varphi: 2^{\feats} \times \feats \rightarrow \mathbb{R}$
% \fif:\reals^{2^{|\feats|}}\rightarrow\reals^{|\feats|}$
that receives the output of a value function and assigns a feature importance score to each feature. For simplicity, we denote the importance of the feature $\feat$ for both the global and the local importance functions as $\fif(\val(z),\feat)$ for $z\in\{x, \dataset\}$, where the actual input for the value function differs between them. An elaborated version of this notation appears in Section~\ref{appendix:framework} of the Appendix. For the data-model discussion, we add several notations: the data itself, $\data$, which is a probability measure over $\dataset \times \target$; $\model$, which is a predictor over $\data$; and $\vald,\valm$ which are indicators for the current mode of evaluation in the value function. 

We say that $\val$ is a valid value function if it satisfies: $\val(\dataset, \emptyset) = 0$, and in the global setting, we further require that monotonicity, i.e. for any subset $\subgroup \subseteq T \Rightarrow\val(\dataset, \subgroup) \leq \val(\dataset, T)$. This reflects the intuition that adding features to a model can not decrease the amount of information regarding the target variable, and thus can not decrease the prediction ability of a model. These two properties imply that $\forall \subgroup \subseteq \feats$, $\val(\dataset,\subgroup)\geq 0$ in the global setting. We do not assume these conditions generally hold for the local setting, implying that $\val(x, \subgroup)$ may be negative for some $x\in \dataset$. Finally, $\expected$ denotes the expected value function with respect to $\data$.

\section{The Local-Global relation}
\label{sec:localglobal}
It is natural to anticipate that a global phenomenon is an aggregate of local phenomena. 
This anticipated consistency can be illustrated intuitively: 
we find it confusing if a model that predicts loan repayment by lenders considers the age of the lenders to be a crucial factor in the global sense, yet at the same time declares that age is not a factor in predicting repayment for any specific lender.
To avoid such scenarios, we require a small set of properties that ensure a meaningful relation between local and global settings in the framework of feature importance.

\subsection{Expected properties}
In this section, we formulate two consistency properties that we require to hold between the local and the global settings. We use these properties to prove the first inconsistency theorem.
\begin{property}[Value Consistency]
\label{property:valueconst}
\label{property: globalrestriction}
$\val$ is \emph{Value Consistent} if 
$$\forall\subgroup \subseteq \feats, ~~
 \val\left(\dataset, \subgroup \right)=\expected\left[\val\left( x,\subgroup \right)\right]$$
        \end{property}
        In property~\ref{property:valueconst}, to establish the relation between the local and the global value functions, the global value of each subset is constrained to be the expectancy taken over the inputs of the local value on this subset.

\begin{property}[Importance  Consistency] \label{property:importanceconst} A tuple $\{\val, \fif\}$ is \emph{Importance Consistent} if
 $$\forall \feat \in \feats,~~ \fif(\val(\dataset),\feat) = \expected[\fif(\val(x),\feat)]$$
     \end{property}

 In property~\ref{property:importanceconst}, to establish the local-global relations of the importance score, a consistency requirement analogous to the one above is made for the feature importance function: the global importance of a feature is the expected value of the local feature importance of this feature.
 
The two properties above define the expected relations between feature importance in local and global settings. We say that a tuple $\{\val, \fif\}$ is local-global consistent to denote that the \emph{Value Consistency} and \emph{Importance Consistency} properties hold.

\subsection{The local-global inconsistency}

We use the MCI function~\cite{catav2021marginal} to demonstrate the discrepancy between local and global settings. This function relies on a pre-defined set of properties which the importance score is expected to maintain. Apparently, the only function that satisfied these properties is the MCI function, defined as follows:
% To establish the local-global inconsistency, we start by introducing the Marginal Contribution Importance (MCI) score, as proposed by~\citet{catav2021marginal}. The MCI score is defined as follows, where $\val$ represents a value function with non-decreasing properties:
$$\text{MCI}(\val, \feat) = \max_{\subgroup \subseteq \feats \setminus \{\feat\}} (\val(\subgroup \cup \{\feat\}) - \val(\subgroup))$$
Remarkably, the MCI score is the only function that uniquely satisfies the MCI properties, detailed in Section~\ref{appendix:mci_properties}. Our analysis leads us to demonstrate the following inconsistency:
\begin{theorem}
\label{theorem:incon1}
properties~\ref{property:valueconst},\ref{property:importanceconst}, and MCI properties do not hold simultaneously.

\end{theorem}

\begin{proofsketch}
    Let $\{\val, \fif\}$ be local-global consistent tuple such that $\val$ is non-decreasing. Assume that MCI properties hold, i.e. $\fif$ is the MCI function. From the local-global consistency, we get that $\forall \feat \in \feats$:
    $$MCI(\expected[(\val(x)],\feat) = MCI(\val(\dataset),\feat) = \expected[MCI(\val(x),\feat)] $$
    This leads to a contradiction since MCI uses the max operator and therefore is a non-linear function of the value function. A proof by counterexample is attached in Section~\ref{appendix:counter_mci}.
\end{proofsketch}
The proof sketch presented here is a simplified version, in which the local importance function and the global importance function are identical. A more detailed version of the proof, which does not assume that, can be found in Section~\ref{appendix:theorem1_detailed}. 

\subsection{Discussion of the local-global relation}
While the global-data setting is defined by Marginal Contribution Importance (MCI), the local-data setting (as the fourth quadrant in Table~\ref{tab:2x2} demonstrates) is much harder to interpret and define. To tackle this issue, the approach adopted in this study was to use MCI's definition of global-data and define the local-global expected relation. However, this led to an inconsistency theorem. The source of inconsistency lies in the different considerations of ambiguous information: MCI ensures that meaningful information is not missed by attributing the maximum contribution to each feature, regardless of the contribution of other correlated features. This differs from methods such as SHAP~\citep{lundberg2017unified}, where contributions are split between correlated features.

\section{The Data-Model relation}

\label{sec:data model}

% An important aspect of explainable  AI is the relation between the explanation of the data and the explanation of the model. 
When explaining data, the focus is on understanding the underlying process generating them; while when explaining the model, the focus is on understanding how the model is making predictions based on the data. However, these settings are intertwined  -- models are often used as proxies by which nature can be explored. In cases where the model predictions are identical to the data, we expect conclusions reached from analyzing the model to hold with regard to the data. Therefore, we expect that the data and model will agree on each feature's importance. Nonetheless, this expected property implies a degenerate case where $\vald \equiv 0$, which implies that for any importance function, the importance score of all features becomes zero, rendering them insignificant.

\subsection{Expected properties}
In this section, we formulate another consistency property, that expresses the expected relations between the data and the model settings. Then, we show that fulfillment of this property, along with known previous results, is only possible in a degenerate case. 
% We will also address problems that arise when feature importance is used for causal inference.

\begin{property}[Data-Model Consistency] 
\label{property:datamodel} 
Let $\model$ be a model that predicts over $\data$. A tuple $\{\data, \model, \val, \fif$\} is Data-Model Consistent if $\forall x,y\sim \data , \model(x) = y$ and $\forall z \in \{x,\dataset\}$ it holds that 
   $$\forall \feat \in \feats,~~ \fif(\vald(z),\feat) = \fif(\valm(z),\feat)$$
         \end{property}
          The \emph{Data-Model Consistency} property~\ref{property:datamodel} states that if a model predicts the target perfectly, then the data and model importance scores of each feature are identical.
   
\subsection{The data-model inconsistency}

We use the SHAP function~\cite{lundberg2017unified} to demonstrate the discrepancy between model and data settings. This function relies on a pre-defined set of properties which the importance score is expected to maintain. Apparently, the only function that satisfied these properties is the SHAP function, defined as follows:
$$\text{SHAP}(\val, \feat) = \sum_{\subgroup\subseteq \feats \setminus \{\feat\}} \frac{|\subgroup|!(|\feats|-|\subgroup|-1)!}{|\feats|!}\cdot (\val(\subgroup\cup \{\feat\}) - \val(\subgroup))$$
Notably, the SHAP score implies additional properties, detailed in Section~\ref{appendix:shap}. This introduction leads us to the following inconsistency:

% We now show a contradiction between the model and the data settings. 
% \begin{theorem}
% \label{theorem:shap}
% Let $\val$ denote a value function. Consider the SHAP function defined as:
% $$\sum_{\subgroup\subseteq \feats \setminus \{\feat\}} \frac{|\subgroup|!(|\feats|-|\subgroup|-1)!}{|\feats|!}\cdot (\val(\subgroup\cup \{\feat\}) - \val(\subgroup))$$
% It is noteworthy that the SHAP function is uniquely characterized as the only scoring mechanism that satisfies the SHAP properties (Detailed definitions of SHAP properties can be found in the Appendix).
% \end{theorem}

\begin{theorem}
\label{theorem:incon2}
If a tuple $\{\data, \model, \val, \fif$\} satisfies Data-Model Consistency~(property~~\ref{property:datamodel}) and SHAP properties, then $\vald\equiv 0$.
\end{theorem}

Our proof is based on a difference between models and the real world. Specifically, when the data contain correlated features, e.g. height measured in centimeters and inches, a model may learn based on only one of the features, resulting in different feature importance scores for each feature in the model. However, in the real world, both features are equally important. 
A detailed proof of the theorem is attached in Section~\ref{appendix:theorem2}.

\subsection{Discussion of the data-model relation} \label{subsec:causality} 
The need to link the data and model settings is not only theoretical. It is motivated by the need to use models to understand how the world works. Feature importance is often used, even if not stated explicitly, as a proxy for causal analysis. Unfortunately, the known limitations of trying to establish causal relations from observational data apply to feature importance too.
The example we used to prove the inconsistency of the data-model often appears in real-world problems. Two features can be highly similar because a common, unobserved variable, caused them, or one of them caused the other. For example, when continuously measuring a variable of interest but only recording its mean and maximum values as observed variables. This problem of lacking information to disentangle the effect of two variables is known as unidentifiability.

To illustrate this in our context, consider two penalized regression models that are trained on two identical features. The first model employs an L1 regularization (lasso regression), and the second model employs an L2 regularization (ridge regression). The predictions of the two models are identical. However, assigning feature importance may lead to different results between the models -  lasso regression will result in assigning all the importance to one of the features, whereas ridge regression will result in assigning equal importance to both features.

\begin{figure}
\begin{center}
\begin{tikzpicture}
\tikzstyle{every node}=[font=\small]
% nodes %
\node[circle, minimum size=0.4cm, text centered, text width=1.8cm, fill=gray!25] (z) {\emph{Smoking (unobserved)}};
\node[circle, right= 3 of z, text height = 0.25cm , text width=1.6cm, text depth=0.12cm, text centered,fill=blue!25] (t) {\emph{Gum}};
\node[circle, right = 7.5 of z, text centered,text width=1.6cm,fill=blue!25, text height = 0.25cm, text depth=0.12cm] (y) {\emph{Earache}};
\node[draw, circle, below = 0.12 of t, text centered, fill=blue!55, text=white, text width=1.6cm, text height = 0.25cm, text depth=0.12cm] (u) {\emph{Cancer}};
% edges %
\draw[->, line width= 1] (z) --  (t);
\draw [->, line width= 1] (y) -- (t);
\draw [->, line width= 1] (z) -- (u);
\draw[->,red, line width= 1,dashed] (y) --node {} (u);
\end{tikzpicture}
 \caption{An example of a directed acyclic graph with a collider variable \emph{Gum}. }\label{fig:collider}
\end{center}
\end{figure}
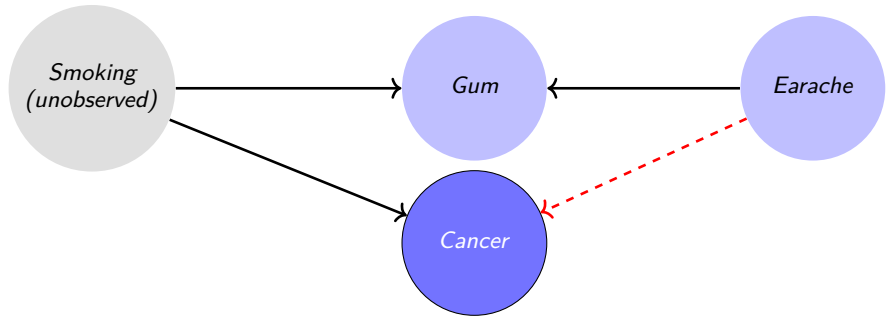

Another situation that may lead to unexpected outcomes from feature importance scores is when a collider (also known as an inverted fork) exists in the data~\citep{pearl2009causality, hernan2010causal}.
For example consider the situation illustrated in Figure~\ref{fig:collider}: Smoking cigarettes (\emph{Smoking}) causes cancer (\emph{Cancer}), but also increases chewing gum consumption (\emph{Gum}). Assume also that doctors recommend people with earaches (\emph{Earache}) to chew gum.
Now, imagine scenarios in which a researcher is developing a model to predict \emph{Cancer} using different subsets of the features \emph{Gum} and \emph{Earache}, but lacks information on \emph{Smoking}. In the first scenario, the researcher uses only the \emph{Earache} feature. Since earache and cancer are independent, any value-based feature importance score will assign zero importance to \emph{Earache}. In the second scenario, where only the \emph{Gum} feature is present, the researcher will conclude that \emph{Gum} is an important feature since it is correlated with \emph{Cancer}.  In the third scenario, where a model that contains both \emph{Earache} and \emph{Gum} is considered, the researcher will infer that \emph{Earache} has non-zero importance.
This results from conditioning on \emph{Gum}, creating an association between \emph{Earache} and \emph{Cancer} due to the presence of a collider. Intuitively, a person who chews gum and does not have an earache is more likely to be a smoker (notice that the smoking feature is unobserved), and hence at high risk of cancer. Therefore, the feature importance score might mislead a na\"ive researcher into thinking that earaches are predictive of cancer and that gum chewing is a cure for the disease.

The situations described here have been studied in the causality literature and there is no recipe for overcoming them that does not involve additional information about the world~\citep{pearl2009causality, hernan2010causal}. 

\section{Conclusion}
\label{sec:conclusion}

In this work, we investigated the possibility to create a unified framework of feature importance scores, by defining their expected properties.
%and the relations between the properties. 
Surprisingly, we found that it is impossible to define feature importance scores that are consistent between different settings. Specifically, the expected consistency between local and global scores contradicts properties of the data-global setting. Furthermore, there is no guarantee that feature importance scores of a model that perfectly predicts the data will reflect the feature importance of the data themselves.

% To overcome these limitations, we proposed an alternative, namely separable sets, that allows assigning scores to sets of features that have inter-set independence with respect to the value function. We showed that a unique maximal partition to separable sets exists and that the demonstrated limitations of feature importance do not apply when using these sets as meta-features. Section~\ref{Sec: Experiments} in the Supplementary Material contains preliminary implementations and experiments that demonstrate the effectiveness of such partitioning. We note that assigning a score to every set of features of size $k$~ has been previously proposed \citep{kumar2021shapley, sundararajan2020shapley}. But in contrast, our partition does not prespecified a set size, nor demands a constant size for each separable set. It relies on the data structure to create sets that yield a consistent feature importance score. Importantly, separable sets also possess the advantage that common feature importance scores converge when applied to them, dismissing the ambiguity of selecting a certain score. 

Our inconsistency result is reminiscent of \citet{kleinberg2002impossibility}, which proves a similar result for clustering. Analogously, we do not argue that we have defined the only possible set of relevant properties for the various settings. We did, however, attempt to define a set of properties that we believe are essential. Yet, even these requirements led to inconsistencies. 
Future research can tackle which further assumptions can be made about feature importance scores, or other explainability methods, that are meaningful and yet can still be consistent. 
%It could be argued that the inconsistencies we identified are invalid, because our assumptions are flawed. 
%For example, the assumptions made by \citet{catav2021marginal} may not be convincing. Or, the assumption that the global score matches the expectancy of the local score is not a proper way to establish consistency between the settings. We nevertheless believe that our work is a step toward a fruitful discussion about what makes an importance score valid.
% Despite these potential pitfalls, we believe that our work contains theoretical developments that integrate previously isolated concepts of feature importance scores.

In the meantime, our results show that feature importance scores should be used cautiously, aligning with recent research that has attempted to measure the quality and usefulness of explainability tools for different applications~\citep{rudin2019we, chang2021interpretable, visani2022statistical}. 
%A possible negative consequence of this study might be attenuating the trust in explainable AI.  However, 
As such, our work tries to promote substantive discussions and accurate definitions of explainability, as previously advocated, for example, by~\citet{lipton2018mythos} and~\citet{kumar2020problems}. Hence, we hope that our work will contribute to stimulating additional research that will result in a solid theoretical foundation for explainable AI.

\bibliography{references.bib}
\bibliographystyle{unsrtnat}

\appendix
\newpage
\label{sec:appendix}

\section{Additional Properties}
In this section, we present the additional properties mentioned in the local-global section and the data-model section. 

\subsection{MCI properties}
\label{appendix:mci_properties}
\citet{catav2021marginal}  introduced the following three properties to define the expected behavior of feature importance scores in the global-data setting:

\begin{property}[Marginal Contribution] \label{property:marginal} A tuple $\{\val, \fif \}$ satisfies the Marginal Contribution property when the importance of a feature is equal to or higher than the increase
in the value function when adding it to all the other features, i.e. $ \fif(\val(\dataset,\feat)) \geq \val(\dataset/,\feats) -
\val(\dataset,\feats\setminus{\feat}))$.
\end{property}
The \emph{Marginal Contribution} property states that the importance of a feature is at least its contribution to the value function when adding the latter to the set of all other features.

\begin{property}[Elimination] \label{property:elimination} A tuple $\{\val, \fif \}$ satisfies the Elimination property when eliminating features from $\feats$ can only decrease the importance of each feature. i.e., if $T \subseteq \feats$ and $\bar\val$ is the value function which is obtained by eliminating $T$ from $\feats$ then 
$$\forall \feat \in \feats \setminus T,~~~ \fif(\val(\dataset,\feat)) \geq  \fif(\bar{\val}(\dataset, \feat))$$
\end{property}
The \emph{Elimination} property states that the importance of a feature does not become smaller when other features are removed from the calculation. The process of elimination is defined as follows:

\newtheorem*{definition*}{Definition}
\begin{definition*}\textbf{(Elimination operation)}
Let $\feats$ be a set of features and $\val$ be a value function. Eliminating the set $T \subset \feats$ creates a new set of features $\feats^\prime = \feats\setminus T$ and a new value function $\val^\prime:2^{\feats^\prime} \rightarrow \reals$ such that $$\forall \subgroup \subseteq \feats^\prime, ~~~~\val^\prime(\subgroup) = \val(\subgroup)$$
\end{definition*}

\begin{property}[Minimalism] \label{property:minimalism} A tuple $\{\val, \fif \}$ satisfies the Minimalism property when for every function $\bar{\fif}:\reals^{2^{\feats}} \rightarrow \reals^{\feats}$ for which properties~\ref{property:marginal} and \ref{property:elimination} hold, then $$\forall\feat \in \feats,~~~ \fif(\val(\dataset, \feat)) \leq \bar{\fif}(\val(\dataset, \feat))$$
\end{property}

The \emph{Minimalism} property states that among all the functions that satisfy properties~\ref{property:marginal} and \ref{property:elimination}, 
the feature importance scoring function should be minimal. 

Using these properties, ~\citet{catav2021marginal} prove the following:
\begin{theorem} \label{theorem:mci}
The MCI feature importance score (see Table~\ref{tab:funcfamily}) is the only score for which the Marginal Contribution property, the Elimination property, and the Minimalism property (Properties \ref{property:marginal},\ref{property:elimination},\ref{property:minimalism}) hold simultaneously.
\end{theorem}

\subsection{SHAP properties}
\label{appendix:shap}
The following properties stem naturally from the SHAP function, which is the only function that satisfies the SHAP properties proposed in~\citet{lundberg2017unified}:

\begin{property}[Triviality]
\label{property:triviality}
A tuple $\{\val, \fif\}$ satisfies the \textit{Triviality Property} if the following conditions hold:
\begin{enumerate}
    \item For all $\subgroup \subseteq \feats$, if $\val(x, \subgroup) \neq 0$, then there exists a feature $f \in \subgroup$ such that $\fif(\val(x), \feat) \neq 0$.
    \item If $\fif(\val(x), \feat) \neq 0$, then there exists a subset $\subgroup \subseteq \feats$ such that $\val(x,\subgroup \cup \{\feat\}) \neq \val(x, \subgroup)$.
\end{enumerate}
The Triviality Property establishes a non-trivial relationship between the value and the importance functions. It requires that if a subset of features has any value, it will be reflected in the importance of at least one feature from this subset. Conversely, it demands that if any feature is important (i.e., has non-zero importance), it must be included in some valuable subset. Notably, if a feature $\feat$ satisfies $\val(x, \subgroup) = \val(x, \subgroup \cup \{\feat\})$ for any subset of features, then $\feat$ has zero importance. 
\end{property}

\begin{property}[Dummy Feature]
\label{property:nullfeat}
Let $\model$ be a model that predicts over $\data$. A tuple $\{\val, \fif\}$ satisfies the \textit{Dummy Feature Property} if, for all $\feat \in \feats$ and for all $x, x^\prime \in \dataset$ such that $x$ differs from $x^\prime$ only by the $\feat$'th feature $\model(x) = \model(x^\prime)$, then 
\begin{equation*}
 \fif(\valm, \feat) = 0
\end{equation*}
The Dummy Feature Property implies that if changing the value of a feature has no effect on a model's output, then the importance of that feature is zero. This property also had been recognized in previous works such as \citet{friedman2004paths} and \citet{sundararajan2017axiomatic}.
\end{property}

\section{Inconsistencies Proofs}
In this section, we present proofs for the inconsistency theorems.

\subsection{Theorem~\ref{theorem:incon1}}
\label{appendix:counter_mci}
Let $\{\val, \fif\}$ be local-global consistent tuple such that $\val$ is non-decreasing. Assume that MCI properties hold, i.e. $\fif$ is the MCI function. From the local-global consistency, we get that $\forall \feat \in \feats$:
    $$MCI(\expected[(\val(x)],\feat) = MCI(\val(\dataset),\feat) = \expected[MCI(\val(x),\feat)] $$
This leads to a contradiction since MCI uses the max operator, and therefore is a non-linear function of the value function.

Now, we aim to demonstrate, by way of a counter-example, that the MCI function is not linear. This will lead to a contradiction between the properties of the data-global, as defined in~\citep{catav2021marginal} setting and $\{\val, \fif \}$ Consistency properties. Formally, we contradict the following equality: 

For any $\val$ which is a valid value function,
\begin{equation}
\label{eq:counter_mci}
 \alpha \cdot MCI(\val(x_0)) + (1-\alpha) \cdot MCI(\val(x_1))  = MCI(\alpha \cdot \val(x_0) + (1-\alpha) \cdot \val(x_1))
\end{equation}

\begin{counterexample}
Let $\dataset$ be a dataset consisting of two samples: $x_0$ and $x_1$, over the feature space $\feats = \{f_0, f_1\}$. We define the value function $\val$ as follows:

\[
\val = \begin{pmatrix}
 & x_0 & x_1 & \dataset \\
\{\emptyset\}: & 0 & 0 & 0 \\
\{\feat_0\}: & 0 & 1 & 1.5 \\
\{\feat_1\}: & 1 & 1 & 1 \\
\{\feat_0,\feat_1\}: & 2 & 1 & 1.5 \\
\end{pmatrix}
\]

Now, let $\alpha = \frac{1}{2}$. We will evaluate the left-hand side of equation~\eqref{eq:counter_mci} and the right-hand side separately.

\textbf{Left-hand side evaluation:}

\[
\begin{aligned}
&\alpha \cdot MCI(\val(x_0)) + (1-\alpha) \cdot MCI(\val(x_1)) \\
&= \frac{1}{2} \cdot MCI\left(\begin{pmatrix} 0 \\ 0 \\ 1 \\ 2 \end{pmatrix}\right) + \frac{1}{2} \cdot MCI\left(\begin{pmatrix} 0 \\ 1 \\ 1 \\ 1 \end{pmatrix}\right) \\
&= \frac{1}{2} \cdot \begin{pmatrix} 1 \\ 1.5 \end{pmatrix} + \frac{1}{2} \cdot \begin{pmatrix} 1 \\ 1 \end{pmatrix} = \begin{pmatrix} 1 \\ 1.25 \end{pmatrix}
\end{aligned}
\]

\textbf{Right-hand side evaluation:}

\[
\begin{aligned}
& MCI\left(\alpha \cdot \val(x_0) + (1-\alpha) \cdot \val(x_1)\right) \\
& = MCI\left(\val(\dataset) \right) \\
&= MCI\left( \begin{pmatrix} 0 \\ 0.5 \\ 1 \\ 1.5 \end{pmatrix}\right) = \begin{pmatrix} 0.5 \\ 1 \end{pmatrix}
\end{aligned}
\]

Hence, we have found a counter-example for equation~\eqref{eq:counter_mci}, which contradicts the claimed linearity of the MCI function. This concludes the proof.
\end{counterexample}

\subsection{Theorem~\ref{theorem:incon2}}
\label{appendix:theorem2}
We establish that any tuple $\{\val, \fif\}$ satisfying the Triviality, Dummy-Feature, and Data-Model Consistency properties (Properties \ref{property:triviality}, \ref{property:nullfeat}, \ref{property:datamodel}) inevitably encounters a scenario where $\vald \equiv 0$. This scenario implies that for any importance function that considers the value, the importance score of all features becomes zero, rendering them insignificant.

\begin{proof}
Let $\data$ be a probability measure over $\dataset$ such that its feature space contains two duplicate features of a random variable, which solely dictate the target. Formally, $\rho \in [0,1]$, $\{f_0, f_1\}\subseteq \feats, \text{ and }\forall x\in \dataset, ~~f_0(x) = f_1(x) = \rho$. The target is defined as $\data(x) =h(\rho)$, where $h$ is some function of $\rho$. Let 
$\model_0, \model_1$ be two models s.t each model focuses on one feature and neglects the other: 
\begin{flalign*}
 \forall i \in\{0,1\},  ~~\model_i(x) = h(\feat_i(x))
\end{flalign*}
Let the tuple $\{\val, \fif\}$ satisfy Triviality, Dummy-Feature and Data-Model Consistency~(properties \ref{property:triviality}, \ref{property:nullfeat}, \ref{property:datamodel}).  By definition, $\model_0, \model_1$ predict the data perfectly, and therefore by the \prop{Data-Model Consistency}, it holds that 
\begin{flalign*}
\forall 
 i \in \{0,1\} \text{ and } \forall \feat \in \feats,~~ \fif (\val^{\data}, \feat) = \fif (\val^{\model_i}, \feat)
\end{flalign*}
 The \prop{Dummy Feature} Axiom implies that 
\begin{flalign*}
\forall 
 i \in \{0,1\}, ~~\fif \left(\val^{\model_i},\feat_{1-i}\right)=0
\end{flalign*}
Combining the last two implies that 
\begin{flalign*}
\forall \feat \in \feats, ~~ \fif \left(\vald,f \right) = 0
\end{flalign*}
By the \prop{Triviality} Axiom, the only value function that satisfies the above is $\vald \equiv 0$.
\end{proof}

\section{Distinguish between the local and global importance functions}
\label{appendix:theorem1_detailed}
In this section, we will reformulate our properties to distinguish between the local and global importance functions. 

\subsection{Framework}
\label{appendix:framework}
Before proceeding, we will provide a more detailed definition of the value and importance functions to ensure precision in describing these functions:

A value function is represented as $\val: (\data \times \{x, \dataset\} \times 2^{\feats}) \rightarrow \reals$. It assigns a scalar to each subset of features $\subgroup \subseteq \feats$. Here, $\val(\data, x, \subgroup)$ signifies the value of a feature subset $\subgroup$ for the local instance $x$, drawn from a probability measure $\data$. Additionally, $\val(\data, \dataset, \subgroup)$ represents the value of the same feature subset over the entire sample space.

On the other hand, a feature importance function is denoted as $\fif:(\{\loc, \glob\} \times 2^{\feats})\rightarrow\reals^{\feats}$. It takes an indicator specifying whether it operates in the local or global context and the output of the value function. This function assigns feature importance scores to individual features. Specifically, $\fif(\loc, \val(\data, x), \feat)$ indicates the importance of feature $\feat$ for the instance $x$, while $\fif(\glob, \val(\data, \dataset), \feat)$ signifies the importance of the same feature across the entire sample space.

Note that the monotonicity property of $\val$ holds in the global setting, but not necessarily in the local setting. For example, consider the case where the prediction target is whether a person has cancer and one of the features is whether the person carries a lighter in their pocket. This feature may be globally important, since it may correlate with smoking. However, it is possible that some people carry lighters but do not smoke, in which case this feature might lead to an erroneous prediction and hence has a negative local contribution. Globally admissible denotes a case where an instance $x$ has only a non-negative contribution. Formally, $\val(\data, x)$ is globally admissible if $\val(\data, x)$ is monotonic non-decreasing and $\val(\data, x, \emptyset)=0$.

\subsection{Expected properties}
\begin{property}[Value Consistency]
$\val$ is \emph{Value Consistent} if for every $\data$ \begin{enumerate}
\item \label{axiom: globalrestriction}$~\forall\subgroup \subseteq \feats, ~~~~  \val\left(\data, \dataset, S\right)=\expected\left[\val\left(\data, x, \subgroup\right)\right]$
\item $\exists x^* \text{ and } \exists\data^*$ such that $\data^*$ is a Dirac measure and $\val(\data^*, x^*) = \val(\data, x)$
\end{enumerate}
\end{property}
To establish the relation between the local and the global value functions, two complementary conditions are required: First, the global value of each subset is constrained to be the expectancy taken over the inputs of the local value on this subset. Second, the local value is constrained to be able to realize the global value.

\begin{property}[Importance  Consistency] A tuple $\{\val, \fif \}$ is \emph{Importance Consistent} for every $\data$
 \begin{enumerate}
     \item $\forall \feat \in \feats, ~~~~ \fif\left(\glob, \val(\data, \dataset), \feat\right) = \expected\left[\fif\left(\loc, \val\left(\data, x\right), \feat\right)\right] $
     \item $\val$ is Value Consistent.
 \end{enumerate}
     \end{property}

 To establish the local-global relations of the importance score, a consistency requirement analogous to the one above is made for the feature importance function: The global importance of a feature is the expected value of the local feature importance of this feature.
 
Consistency implies a commutative diagram, which is presented in Figure~\ref{tab:locglob}. 

\usetikzlibrary {graphs,quotes,shapes.geometric}
\begin{figure}[t]
\centering
{\caption{{\bf consistency diagram:} In local-global consistency the global value is the expectation of the local values, while the global importance is the expected value of local importances. }% caption for whole figure
{%
\label{tab:locglob}
{
\begin{tikzpicture}
\graph { 
  a/$\val(x)$ ->  b/$\fifdiagloc$[xshift=1cm],
  a -> ["$\expected$"] c/$\val(\dataset)$[yshift=-1cm],  
  c -> d/$\fifdiag$[xshift=2cm],
  b ->["$\expected$"] d;
};
\end{tikzpicture}
}
}\qquad % space out the images a bit
}
\end{figure}

\usetikzlibrary {graphs,quotes,shapes.geometric}
\begin{figure}[t]
\centering
{\caption{{\bf consistency diagram:} In local-global consistency the global value is the expectation of the local values, while the global importance is the expected value of local importances. }% caption for whole figure
{%
\label{tab:locglob}
{
\begin{tikzpicture}
\graph { 
a/$\data$ -> b/$\vald$[xshift=1cm] -> c/$\fif(\vald(z))$[xshift=2cm],
a -> ["$\equiv$"] d/$\model$[yshift=-1cm],
d -> e/$\valm$[xshift=2cm] -> f/$\fif(\valm(z))$[xshift=3cm],
c -> ["$\equiv$"]f;
};
\end{tikzpicture}
}
}\qquad % space out the images a bit
}
\end{figure}

% Lemma~\ref{lemma:the local is the global} states that for every value which is globally admissible, the global and local feature importance functions are identical. Lemma~\ref{lemma:linear} builds on Lemma~\ref{lemma:the local is the global} to state that under local-global consistency (i.e Axioms~\ref{axiom:valueconst}, \ref{axiom:importanceconst} hold) the global importance function has linear properties. This contradicts Theorem~\ref{theorem:mci} since MCI uses the max operator, and by that not linear.

\subsection{Detailed proof for Theorem~\ref{theorem:incon1}}

The proof uses the following two lemmas. Combining these lemmas implies that $\fif$ is a linear function, and the rest of the proof is identical to the abbreviated version that appears above.  

\begin{lemma} \label{lemma:the local is the global}
Let $\data$ be a probability measure over $\dataset$. If $\{\val, \fif\}$ is importance consistent and $x\in \dataset$ such that $\val(\data, x)$ is globally admissible, then $$\fif(\glob, \val(\data,x)) = \fif(\loc, \val(\data, x))$$
\end{lemma}

\begin{proof}[of Lemma~\ref{lemma:the local is the global}]
Let $\{\val, \fif\}$ be important consistent tuple. Let $\data$ be a probability measure over $\dataset$ and let $x\in\dataset$ be a globally admissible instance. Denote $\data^\prime$ as the corresponding probability measure, i.e $\val(\data, x) = \val(\data^\prime, \dataset^\prime)$. By the \prop{Value Consistency} property there exist a Dirac measure $\data^*$ such that $\val(\data^\prime, \dataset^\prime) = \val(\data^*, x^*)$. Hence,
\begin{align}
\fif\left(\glob, \val(\data, x)\right) 
& = \fif\left(\glob, \val(\data^\prime, \dataset^\prime)\right) \\
& = \fif\left(\glob, \val(\data^*, \dataset^*)\right) \label{eq:D star}\\
& = \fif\left(\loc, \val(\data^*, x^*)\right) \label{eq: D star Dirac}\\
& = \fif\left(\loc, \val(\data^\prime, x^\prime)\right) \\
& = \fif\left(\loc, \val(\data, x)\right) 
\end{align}

where~(\ref{eq:D star}) is from the global admissibility of $\val(\data, x)$, (\ref{eq: D star Dirac}) follows from the consistency and from the fact that $\data^*$ is Dirac, and the following equations follow from the definition of $\data^\prime$.
\end{proof}

\begin{lemma} \label{lemma:linear}
Let $\data$ be a probability measure over $\dataset$ and let $\val$ be a monotonic non-decreasing value function (i.e. $\forall x \in \dataset, ~~\val(\data, x)$ is globally admissible). If $\{\val, \fif\}$ is local-global consistent then 
$$\fif\left(\glob, \expected[(\val(\data, x)]\right))=\expected[\fif\left(\glob, \val(\data, x)\right) ]$$
\end{lemma} 
\label{appendix:lemmas}

\begin{proof}[of Lemma~\ref{lemma:linear}]
Let $\{\val, \fif\}$ be a local-global consistent tuple and let $\data$ be such that $\val(\data, x)$ is globally admissible for every $x$ in the support of $\data$.
% \footnote{For the sake of conciseness we use the symbol $\data$ both as a probability measure on $\dataset\times\target$ and its projection on $\dataset$.}
Therefore,

\begin{align}
 &\expected[ \fif\left(\glob, \val(\data, x))\right)]\\
& \label{eq:uselemma} 
 =\expected[ \fif\left(\loc, \val(\data, x))\right)]\\
&\label{eq:importance}= \fif\left(\glob, \val(\data, \dataset) \right)\\
&\label{eq:value}= \fif\left(\glob, \expected[\val(\data, x)]\right) 
\end{align}
where~(\ref{eq:uselemma}) is valid by Lemma~\ref{lemma:the local is the global},and (\ref{eq:importance}) and (\ref{eq:value}) are valid by the importance consistency~\ref{property:importanceconst}.
\end{proof}

% $\val$ is the value function that assigns a scalar to each subset of features $S \subseteq \feats$. \\
% $\varphi$ is the importance function that assigns a score to each feature based on its contribution. \\
% $\mathcal{D} / \mathcal{M}$ indicates the source for the value evaluation process.
% $$\varphi(\nu^\mathcal{M}(x,f))$$
% $$\varphi(\nu^\mathcal{M}(X,f))$$
% $$\varphi(\nu^\mathcal{D}(x,f))$$
% $$\varphi(\nu^\mathcal{D}(X,f))$$
\end{document}